\documentclass[letterpaper, 10 pt, conference]{ieeeconf}  
\usepackage{graphicx, verbatim}
\usepackage{ifthen,latexsym,amssymb,amsmath}
\usepackage{hyperref}
\usepackage[draft=true]{minted}
\usepackage{color,soul}

\IEEEoverridecommandlockouts                              

\title{\LARGE \bf
Open Problems in Robotic Anomaly Detection
}

\author{Ritwik Gupta$^{1}$, Zachary T. Kurtz$^{1}$, Sebastian Scherer$^{2}$, and Jonathon M. Smereka$^{3}$
\thanks{*This work was sponsored by the U.S. Army Tank Automotive Research, Development, and Engineering Center (TARDEC)}%
\thanks{$^{1}$Ritwik Gupta and Zachary T. Kurtz are with the Software Engineering Institute, Carnegie Mellon University, 4500 Fifth Avenue, Pittsburgh, PA 15213, USA
        {\tt\small \{rgupta, ztkurtz\}@sei.cmu.edu}}%
\thanks{$^{2}$Sebastian Scherer is with the Robotics Institute, Carnegie Mellon University, 5000 Forbes Avenue, Pittsburgh 15213, PA, USA
        {\tt\small basti@andrew.cmu.edu}}%
\thanks{$^{3}$Jonathon M. Smereka is with U.S. Army TARDEC, Warren, MI 48397, USA
	{\tt\small jonathon.m.smereka.civ@mail.mil}}%
}

\begin{document}

\maketitle
\thispagestyle{empty}
\pagestyle{empty}

\begin{abstract}
Failures in robotics can have disastrous consequences that worsen rapidly over time. This, the ability to rely on robotic systems, depends on our ability to monitor them and intercede when necessary, manually or autonomously. Prior work in this area surveys intrusion detection and security challenges in robotics, but a discussion of the more general anomaly detection problems is lacking. As such, we provide a brief insight-focused discussion and frameworks of thought on some compelling open problems with anomaly detection in robotic systems. Namely, we discuss non-malicious faults, invalid data, intentional anomalous behavior, hierarchical anomaly detection, distribution of computation, and anomaly correction on the fly. We demonstrate the need for additional work in these areas by providing a case study which examines the limitations of implementing a basic anomaly detection (AD) system in the Robot Operating System (ROS) 2 middleware. Showing that if even supporting a basic system is a significant hurdle, the path to more complex and advanced AD systems is even more problematic.  We discuss these ROS 2 platform limitations to support solutions in robotic anomaly detection and provide recommendations to address the issues discovered.
\end{abstract}

\section{Introduction}
Anomaly detection (AD) is an increasingly important area of study in the field of robotics as robotic systems tend towards higher levels of autonomy. Being able to predict, identify, and correct these anomalies is critical, especially when the robotic systems can have a direct or indirect impact on human life. Unfortunately, while all versions of anomaly detection seek to identify things that are anomalous, there is still considerable variation in precisely what this means:
\begin{enumerate}
\item \underline{Extreme}: The point lies above a threshold $t$.
\item \underline{Isolated}: In some metric space, the distance to other points is greater than $t$ except for at most $n$ of other very nearby points (a point at the center of a highly bimodal distribution can be isolated and not extreme).
\item \underline{Abnormal} (or \emph{inconsistent with a trusted model}): As an example, an auditor keeps track of the ratio of total income to total taxes paid for a collection of organizations. One organization is far larger than the others, with income and taxes being both extremely high. However, the ratio of taxes to income for this large organization is comparable to the ratio for smaller organizations, and the auditor considers it normal. Thus, a point can be both extreme and/or isolated and yet still fail to be abnormal.
\end{enumerate}
The differences between the senses above are conceptually superficial. For any space containing an isolated point, there exists a simple transformation of the space that results in the isolated point becoming an extreme value. Similarly, the size (in terms of income and taxes) of an organization is really a distraction if the ratio of income to taxes is what matters, so why not just talk about that ratio? Unfortunately, while these kinds of conceptual connections between competing notions of anomalousness are trivial for simple examples, they become less trivial as the dimension of the space grows. 

The anomaly detection task is especially challenging when we are asked to treat the data as a black box, with no \textit{a priori} insight into what is ``normal". A general-purpose anomaly detection algorithm will require considerable sophistication to automatically notice the relationship between income and taxes without any prior knowledge of finance. Accordingly, varying techniques of anomaly detection in robotic monitoring focus on predefined relationships of what is a ``normal range'' of operation \cite{sifferAnomalyDetectionStreams2017, scheinBayesianPoissonTensor2015, friedlandClassifierAdjustedDensityEstimation2014, bruns-smithCyberSecurityMultidimensional2016, speakmanPenalizedFastSubset2016}, however, as we show in this work, there are still several open problems in robotic anomaly detection that significantly degrade the assumption of being able to define that ``normal range''.

Finally, we demonstrate the need for additional work in these areas by providing a case study which examines the limitations of implementing a basic anomaly detection (AD) system in the Robot Operating System (ROS) 2 middleware \cite{ROS2Design}, which is an attempt to revise and improve many engineering decisions from the ROS 1 platform \cite{ROS2009}. ROS has often been difficult to work with and requires specific engineering guidelines which are not conducive to real-time anomaly detection.  Accordingly, we draw the conclusion that if even supporting a basic system is a significant hurdle, the path to more complex and advanced AD systems is even more problematic.  We discuss these ROS 2 platform limitations to support solutions in robotic anomaly detection and provide recommendations to address the issues discovered.


\section{Open problems with regards to robotic AD systems}\label{open_problems}

\subsection{Non-malicious faults present many false alarms.}\label{open_problems_nonmalicious}

False positives and false negatives have been well studied in AD and intrusion detection systems \cite{patchaOverviewAnomalyDetection2007,grillReducingFalsePositives2017,timmStrategiesReduceFalse2001}. It is a long-held belief that an anomaly means a failure of a system directly. However, \textit{not all anomalies represent failures}. A robot can behave anomalously frequently without ever failing, resulting in a large amount of false alarms that, in turn, results in operator overload and cognitive burden. Alarmingly, high false-alarm rates lead to lowered operator trust in robots \cite{chenEffectsImperfectAutomation2009}.

A big issue with false alarms is that being able to handle them means that we have \textit{a priori} information that a given alarm is a false positive, meaning the alarm should not have occurred in the first place. There are not many methods to eliminate false alarms beyond creating methods with better sensitivity and specificity. The current dominant method to handle operator burden due to these false alarms relies on alarm rate thresholding \cite{chandolaAnomalyDetectionSurvey2009}. Tweaking the threshold at which an alarm persists on the operator's alert panel, in practice, allows many false alarms to be handled at the cost of tuning this subjective parameter and possibly suppressing one-off true alarms. Methods to dynamically change the threshold exist \cite{everettSupervisedAutonomousSecurity1988}, but they are brittle and are still prone to suppressing true alarms. A proposed method of handling false alarms is by using a pseudo human-in-the-loop (HITL) approach, in which all alerts are initially presented to the operator. As the operator dismisses alerts, a simple model can learn the operator's preferences and subsequently use the feedback to suppress false alarms, or alarms that the operator deems unnecessary. Approaches like these have been explored in security contexts \cite{cranorFrameworkReasoningHuman2008} but have yet to be seriously applied in robotics contexts.

It is infeasible to handle non-malicious faults as it is impossible to know ahead of time which faults are non-malicious. Since the idea of which alert is alarming is subjective based on the situation and the operator, HITL approaches present a dynamic and robust way to handle alerts related to non-malicious faults. HITL systems need further interest and development with regard to AD systems.

\subsection{When is invalid data anomalous, and when is it not?}\label{open_problems_invalid}

Conceptually, one can separate the data ingestion pipeline into three parts: 1) the sensor, which is the source of the data, 2) the data processing pipeline, which does the initial data transformation/cleaning, and 3) the data interpretation pipeline, which provides meaning to the data in the context of the overall robotic system (often the data processing and interpretation pipelines are merged, but a distinction still exists). It is often the case that the beginning stage of the pipeline (the sensor) may be operating without anomalies but returning data that is ``invalid" from the standpoint of the remaining parts of the pipeline that is interpreting the data or monitoring a user. In this case, it is hard to say that invalid data is ``anomalous."

There are two schools of thought on this matter:

\begin{enumerate}
\item The data is never anomalous, but the interpretations are. 
\item The data can be flawed given a static interpretation framework.
\end{enumerate}

The former, data-purist school of thought is compelling because it conditions us to replace human senses and intuition with robotic ones such that the available sensors and all their data (invalid or not) are the ground truth. Similar to an instance where a human believes they saw a UFO, the data itself (the light collected by the eyes) is always valid, but frameworks of interpretation (the way the brain processes the light collected by the eyes) may be incorrect or behave in anomalous fashions. In this model, the AD burden falls to the data interpretation pipeline.

The latter school of thought is arguably more practical because it allows the engineer to define limits on the normalcy of data. Imagine a use case where a LiDAR sensor is collecting point cloud data to map its surrounding region. A splash of water from a passing car covers the sensor for a few seconds. In this perspective, the data from the sensor needs to be ignored as valid data because the sensor's operations are compromised. The AD burden in thsi scenario lies with the data processing pipeline.

Overall, \textit{invalid data cannot be directly correlated with anomalous behavior}, though a correlation certainly exists. It is essential when building AD systems to take a softer stance towards invalid data and provide methods to filter it out without triggering alarms about anomalous behavior. Methods to filter invalid data vary \cite{murphyHandlingSensingFailures1999, goncalvesSystemsMethodsFiltering2007, pignatiPreestimationFilteringProcess2014}, and there is no one right way to approach this problem. Robust sensor and domain-specific filtering methods need to be generalizable to handle unforeseen edge cases.

It is also important to consider that anomalous states are, again, often defined as some deviation from a typical behavior trend. This ``normalcy" carries with it the idea that there are base assumptions about operation, environment, and intended behavior. That is, if there is some function $F$ that defines normal behavior, it is only valid under a set of assumptions $A$. Current robotic AD systems do not monitor $A$, even if $A$ is prone to change. For example, a robotic system that is meant to only operate on flat, 2D surfaces must suddenly operate on a 2D plane that exists on a slope. The base assumptions for this robotic system have changed, and this state should be detected as an anomaly. While this change in base assumptions may have an effect on the overall operation of the robot and still can be detected via the classical, deviation from $F$ definition of anomalous behavior, it is sometimes more effective to capture changes in $A$.

By neglecting to account for the ever-changing set of assumptions about the environment, current robotic AD systems are severely lacking in their ability to handle the dynamic environments that current robotics research focuses on. In order to reason about changes in $A$, we must reason about the environment with a non-static framework of ``normalcy." As robotic systems progress further into autonomous operation in unconstrained environments, future AD systems will need to capture changes in $A$, which will require new ideas and research into unreliable surroundings.

\subsection{Intentional anomalous behavior and emergency stops}

Robotics failures are far from monolithic. Some failures are internal, as when a part breaks or when an algorithm fails to anticipate a logical outcome. Some failures are external, when the power grid stops powering the robot or an operator drives the robot over a cliff. And some failures have no simple blame prescription. An operator and a robot may jointly confuse each other; a robot may stumble into an environment for which it was not designed to be successful; or an operator may push the robot to the edge of its limits in a completely rational act of desperation.

In order to characterize intentional anomalous behavior and how to apply emergency stops in particularly alarming cases, we must return to forming a definition of normal operating conditions with respect to the explicit physical bounds of the robotic system. Examples include the amount of weight an arm can lift, degrees of rotation, and the speed at which something can operate. We can imagine these bounds as a closed curve in some operation envelope. As a safety measure, most robots are rated to perform outside their normal operating conditions.

\begin{figure}
	\centering
	\includegraphics[width=3in]{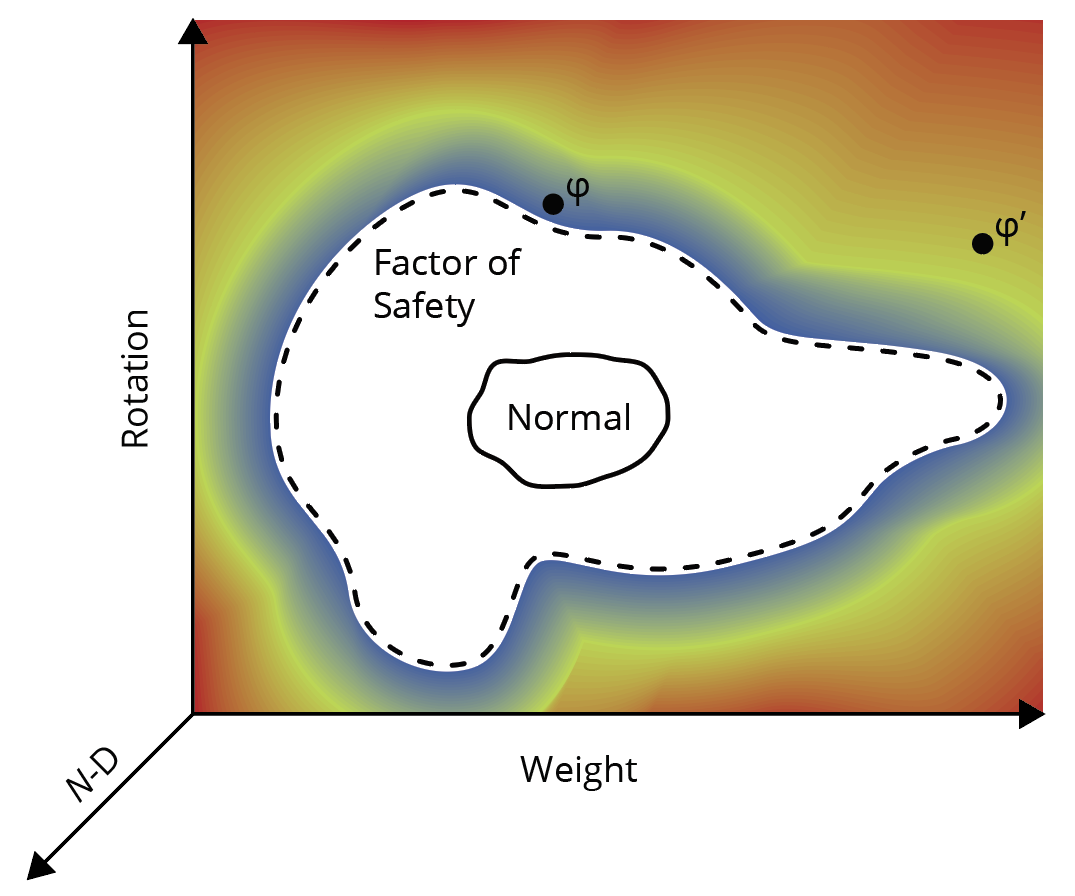}
	\caption{Example of some arbitrary operation envelope $OE$}
	\label{img_normal_operation}
\end{figure}

For some $N$-dimensional set of all possible operating conditions $OE$, where each dimension represents an operating attribute (weight, rotation, etc.), we have $N \subseteq FOS \subseteq OE$, where $N$ is the subset of operating conditions that fall within normal operating bounds and $FOS$ represents the subset of operating conditions that take into account some \textit{Factor of Safety} (FoS). $FOS^\complement = FOS \setminus OE$ represents the set of operating conditions that are outside of the FoS and are not accounted for by the engineers of the system.

Given some state $\phi \in OE$, when can one say that it represents anomalous physical behavior? Certainly if $\phi \in N$, then $\phi$ is not anomalous. On the other hand, if $\phi$ is very much outside the bounds of $FOS$, then it is anomalous. There exists this gray area in between. When $\phi$ is outside of $FOS$ by some small margin, it may not necessarily be anomalous, yet when the margin is appreciable then the probability that the state is anomalous increases. An example here is redlining a car. Redlining by itself may not be anomalous behavior; however, the probability of identifying that state as anomalous increases as we maintain the redline.

Assume that there exists a \textit{risk surface} outside the bounds of $FOS$. The risk surface defines how ``risky" any $\phi$ is as it moves further from $FOS$. In the case of robotics, it is necessarily true that the steepness of the risk surface and the ``distance" (for some abstract, non-continuous definition of distance) away from $FOS$ is directly correlated. This risk surface is more often than not unknown since $OE$ is generally high-dimensional and many configurations are untested. Therefore, while some of the risk surface may be defined via explicit testing, a large portion of it must be learned or inferred. A \textit{risk model} then defines a function over the risk surface that quantifies the amount of risk present in any given $\phi$. It is important to state that a risk model may either be explicitly defined via rigorous testing if $OE$ is small, but will have to be learned if $OE$ is of any sizable cardinality (as in the case of autonomous driving, for example).

Once we have a quantified view of the ``riskiness" of a state, we can then create an AD system by integrating time and count components to see how long or how often a risky state is maintained. Using either thresholding or a learned cut-off, it is then feasible --- and most importantly, explainable --- to trigger emergency stops at relevant times in the robotic system's execution.

Intentional anomalous behavior is currently not well understood. As such, autonomous emergency stops are difficult to calculate and execute. It is imperative to define risk models ahead of time that can give scales of confidence with which to implement emergency stops; defining risk models in this is manner currently under-explored.

\subsection{AD systems built around hierarchies of systems with shared functionality} \label{open_problems_hierarchy}

Common systems fail in similar fashions. For example, autonomous room cleaners have a shared mode of failure in that they may localize incorrectly, or different types of load bearing arms get stressed on common joints. Therefore, instead of building AD systems that are sensitive to each sensor and actuator, it is possible to create an ontology/hierarchy of sensors and actuators to simplify the AD burden. Such approaches have been explored \cite{xiongHierarchicalPM2011} and demonstrate significant improvement in AD capability, but rely on explicitly defined groupings which may not exist in a robotics use-case. Most systems in a robotic platform are not fundamentally unique and failures on these systems do not need to be finely monitored. By clustering their behavior together and focusing on a subset of their messages features that are shared across these systems, we can reduce the number of topics a robotic AD system must focus on. This reduces computation load and operator burden, as tailored alerts for each system are no longer provided for non-critical systems.

Robotic systems can be described via their information-processing aspects terms in of a graph structure:
\begin{itemize}
	\item a collection of $k$ nodes $V = \{v_1, ..., v_k\}$, where some nodes are connected by directed edges $E = \{(v_i, v_j)\}$ variously representing physical anchoring, energy flow, or information flow of various kinds,
	\item the graph is defined as $G = (V, E)$,
	\item and all nodes are loyal to some pre-defined objective (at least implicitly, by design).
\end{itemize}
However, in order to discuss shared functionality, communication graphs are not an ideal representation. An alternate definition of a robotic system that may be more conducive is as follows:
\begin{itemize}
	\item a collection of $k$ nodes $V = \{v_1, ..., v_k\}$, where nodes are connected by directed edges $E = \{(v_i, v_j)\}$ representing one-way communication channels,
	\item the graph is defined as $G = (V, E)$,
	\item nodes can be grouped in the form of $\{v_x | f(v_x)\}$  $\exists v_x \in C$, where $f(x)$ represents a predicate function that returns true if $v_x$ has a certain functionality, and $C$ represents the overall set of all groups in the robotic system,
	\item and $v_x$ is a member of only one subset of $C$
\end{itemize}

It is easy to imagine a sub-component of a robotic system as a monolith; if we can abstract away the finer details (e.g., imagine the combination of a robotic ``wrist" and ``elbow" as just a robotic ``arm") then we can reduce cognitive and computational burden. We call this ability to view things as a simple hierarchy \textit{composability} --- the ability to compose items together into a higher-level merger of objects. Let the behavior of nodes $V$ be denoted by $\boldsymbol{B} = [b_1, ..., b_k]$, where $|\boldsymbol{B}| = |V|$. Let there also be a vector of constants $\boldsymbol{\Phi} = [\alpha_1, ..., \alpha_k]$, $|\boldsymbol{\Phi}| = |\boldsymbol{B}|$. \textit{Linear composability} is then defined by $\boldsymbol{\Phi}^T \cdot \boldsymbol{B} = \alpha_1b_1 + ... + \alpha_kb_k$. This necessarily means that the behavior of each node is independent from the behavior of other nodes, which means that this composition is \textit{decomposable}.

Attributes of nodes that are linearly composable are well-suited to hierarchical treatment. However, most robotic attributes are not inherently decomposable. For example, the dynamics of a robotic ``arm" is a \textit{non-linear composition} of the ``wrist" and ``elbow". The dynamics on the wrist interact with the elbow in a dependent fashion. Thus, there is sizable information lost when treating both components as one hierarchy. However, attributes such as individual CPU load can easily be treated as one hierarchy for AD purposes.

There are many gains to be had by abstracting the AD burden into hierarchies of systems, but there is not a clear understanding of what systems can be treated as a group. We present a minimal framework of thought with which to tackle this problem; however, there is not sufficient literature in this area to make definitive conclusions about the merit of this approach.

\subsection{Distribution of computation across hosts} \label{open_problems_distribution}

Anomaly detection algorithms sometimes require complex computation that may not be possible on many robot platforms due to limited computational resources. With the prevalence of network connected devices, it is possible to distribute computation to nodes that possess sufficient resources to run computation and send results back to the robot platform.

Distributed computation will be the cornerstone of robotics in the future. With the large amount of sensor, log, and systems data that can be collected and generated by robotic systems, it will become infeasible to provide enough computation on one system to process all the data in a timely manner, especially with regard to AD systems.

There are many well-established methods to distribute computation. To this point, we will discuss three main architectures we believe are especially pertinent to robotics: peer-based, hub-and-spoke, and local reduction with a hub.

\begin{figure}[h]
	\centering
	\includegraphics[width=1.6in]{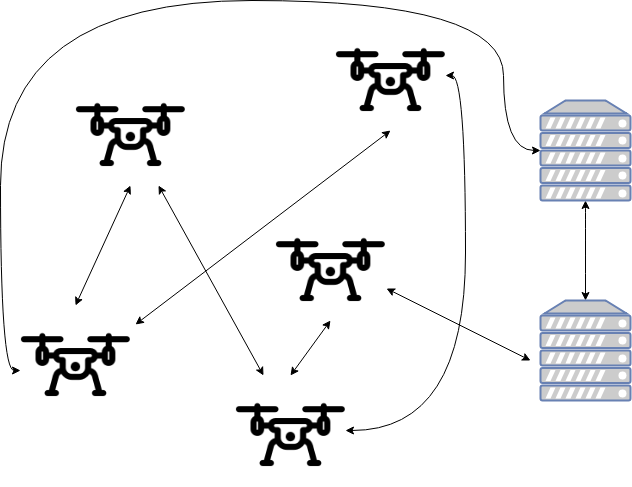}
	\caption{Notional peer-to-peer architecture}
	\label{img_peer_to_peer}
\end{figure}

Peer-based computation is a model that relies on all peers in a network performing computation together in order to quickly solve a given task. Many such models have been explored \cite{milojicicPeertoPeerComputing, bicksonPeertoPeerSecureMultiparty2008, kingSecureScalableComputation2006}, but the central idea is that some function $F$ can be passed to all nodes in a system, and then some data $X$ can be processed by all nodes with results broadcast to all peers. Additional overhead is optional if results need to be merged locally. Alternatively, nodes can be assigned unique functions to compute and a pipeline can be created across nodes, such that no one system is responsible for all the work. In robotic AD systems, the peer-to-peer approach is viable as the AD computation could feasibly be passed around to nodes that are not actively utilizing their computational resources with the benefits being relayed to the entire system. Furthermore, a single node going out of contention does not harm the overall system. This approach has been explored \cite{bijayendrayodhinPeerAgent2002, parkerRandomPeer2007}, but present many limitations, such as network flow problems associated with a slow peer, protocol issues, and more.

\begin{figure}[h]
	\centering
	\includegraphics[width=1.6in]{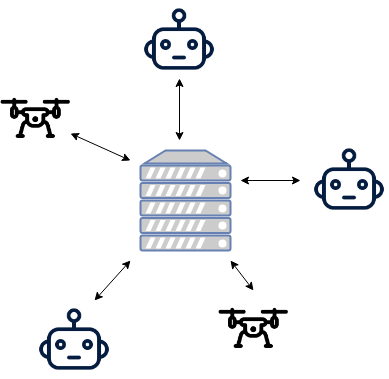}
	\caption{Notional hub-and-spoke architecture}
	\label{img_hub_spoke}
\end{figure}

Hub-and-spoke (often called server/client architecture) models are simple: all computation is offloaded to a central server. The server receives the data, processes it, and returns results to all the ``spokes" (nodes), \textit{ad infinitum}. This removes the burden of computation from the robotic system entirely, which means that robots can be deployed with less computational power than otherwise necessary. This is ideal when AD systems have to deal with massive amounts of data. This idea has been widely embraced by the field of ``cloud robotics", for which \cite{duRobotCloudCenter2011} and \cite{huCloudRobotics2012} give a good background. This method, however, introduces multiple modes of failure, such as loss of autonomy when there is a loss of signal, computation server failures, and more.

\begin{figure}[h]
	\centering
	\includegraphics[width=1.6in]{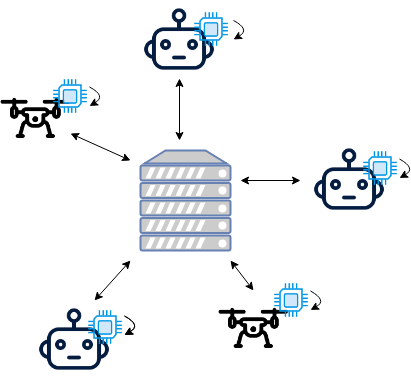}
	\caption{Notional local reduction with hub-and-spoke architecture}
	\label{img_hub_spoke_local}
\end{figure}

An extension of the hub-and-spoke architecture to reduce the computation load on the central hub is to introduce the concept of local reduction. Each robotic system or component can perform local computation within the bounds of its computational resources by itself, and then pass intermediate results to the hub, where it can process further with lower computational load. This is an exciting model, as robots can already process their data comfortably, so an AD system can still process massive amounts of data with less computational load across the entire system.

The central problem that needs to be addressed is that of \textit{autonomy vs. computation}. By relying on systems outside of the robotic system itself, the robot loses autonomy. It now must account for communication failures, node dropouts, and countless other errors that could cause it to no longer have the ability to perform some computation. On the other hand, by allowing for distributed computation, the robotic system gains the ability to process more data than it can by itself. These trade-offs are critical and may vastly differ based on use case. \textit{Graceful degradation} of communication abilities or node failures need to be studied in detail, as the fundamental operations of a robotic system must be maintained even when all distributed computation loads go awry.

For many existing and emerging robotic use cases, such as manufacturing or self-driving vehicles, distributed computation of AD computations will allow for the greatest flexibility in anomaly detection, correction, and recovery procedures.

%
%
%
%

\subsection{Fixing anomalies on the fly}\label{open_problems_fixing}

Once anomalies have been identified, many systems do not take much action beyond providing the operator or the log with an alert that can be manually corrected, if the operator deems it fit. However, with autonomous systems, it is essential that automatic correction of errors and anomalous behavior be integrated as a component or extension of an AD system. Flight control systems include fault-tolerance and error correction as one of their core components \cite{springerFlightControl2010}

A way of introducing fault tolerance into systems is to run multiple versions of the same system simultaneously and switch between the systems if anomalous behavior is detected. This method has been explored for visual odometry \cite{holtzLearningContextDependentSwitching2016} and in other fields, such as optimization and distributed computing \cite{karakusRedundancyTechniquesStraggler2018}. This method is successful but computationally inefficient, as computation would increase with a complexity of $\Omega(N)$ with respect to the number of redundant systems.

Non-blocking snapshots \cite{chandyDistributedSnapshotsDetermining1985} of critical systems during periods in which no anomalous behavior is occurring constitute an alternative way of fixing anomalous behavior or errors within various systems. By reverting to a snapshot that is known to work, the system can recover from anomalous behavior, and may even avoid falling back to its anomalous state given that the initial anomaly was caused via non-deterministic causes. The downside to this is that the system may be stuck in a snapshot-recovery cycle if there is a deterministic and reproducible cause that the system is facing.

A middle ground to non-blocking snapshots and running multiple copies of a system is the concept of shadow computing and shadow replication \cite{millsShadowComputingEnergyaware2014}, in which a process is duplicated with \textit{shadow processes}. These shadow processes are an exact replica of the main process but are executed at a lower quality of service (QoS). In robotics, mission-critical software and hardware drivers can be shadow replicated, and upon the detection of anomalous behavior due to a non-deterministic fault, a healthy shadow process can be substituted. The anomalous process can then be shadowed and restarted as needed.

Robotic systems can also learn to correct anomalous behavior and incorporate error correction into various algorithms in the overall system. This approach is explored in the context of planning \cite{mendozaRegionsInaccurateModeling2017}, map learning \cite{engelsonErrorCorrectionMobile1992}, and many other application domains \cite{mojaradClusterBased2016, hongminRecoveringExternal2017, guoNonlinearUnknown2018}. Better error correction is a topic of active research in many fields \cite{stumpeKeplerErrorCorrection2012, santiniAnomalousPixelCorrection2014, bushThermalEffects2000}. A truly global method of error correction is out of scope given current technology since the combined dimensionality of all systems in a robot are too vast for learning algorithms to be able to map.

In order to properly address these concerns, it is imperative for anomaly correction to be seen as a basic research task rather than an applied one. Furthermore, borrowing existent research from other fields is critical, as a lot of fundamental work already exists and needs to be adapted for the robotics-specific domain. Finally, AD correction also needs to be quantified in terms of operational risk, i.e., it is currently not well understood how much the \textit{lack} of anomaly correction affects missions in terms of cost and resources.


\section{Monitoring our own robot: a case study}

We implemented a basic AD system in ROS 2 to demonstrate the need for more work in examining robotic AD systems. Due to the popularity and decisions behind the ground-up redesign of ROS 2 \cite{ROS2Design}, we show that basic architectural constraints to implementing a real-time AD system reinforce the conclusion that supporting future AD systems is currently not possible without changes to the system.

The platform we chose for implementation is the Turtlebot 2\footnote{https://www.turtlebot.com/turtlebot2/}, which is a low-cost, open-source robot that provides a framework on which to build. Traditionally, the Turtlebot 2 runs on ROS. Efforts have been made\footnote{https://github.com/ros2/turtlebot2\_demo} to run ROS 2 on the base, but they largely rely on the ROS 1 bridge. Nevertheless, it provides us a target to start developing new ROS 2 technologies.

\subsection{Control}\label{our_control}

Tele-operation (tele-op) is a common way to control robots in many environments. In order to support full tele-op for the Turtlebot on ROS 2, we created a ROS 2 node that allows an Xbox controller to be used to control all degrees of motion for the Kobuki base.
\begin{itemize}
	\item This is a complete ROS 2 \verb|rclpy| implementation and does not rely on the ROS 1 bridge.
	\item Both publishing and subscription occur in the same node to listen to Joy messages and send messages to \verb|/cmd_vel|.
\end{itemize}


\subsection{Capturing Messages}\label{our_messages}

We explored two methods to capture the messages published on topics in the ROS system. Both methods have their upsides and downsides. For the sake of modularity, we focus only on the upsides here and capture the limitations later.
\begin{enumerate}
	\item We created an \verb|rclpy| ROS 2 node that would automatically subscribe to all the topics in the system and write their messages to disk.
	\begin{itemize}
		\item We can debug messages in real-time with existent Python tooling.
		\item New topics can be subscribed to on the fly without modifying the ROS environment.
		\item The behavior of the topic collector can be changed on the fly by publishing messages to it.
		\item Can be extended to perform actions on the ROS 2 system in real-time.
	\end{itemize}
	\item Using \verb|rosbag| over the ROS 1 bridge.
	\begin{itemize}
		\item Use existent ROS infrastructure without having to reinvent the wheel.
		\item Easy tooling to analyze and simulate \verb|rosbag| files.
	\end{itemize}
\end{enumerate}
Both methods give us files that we can interpret as a chronological feed of the robot's operation, which we can analyze and perform actions on. Of the two, Method 1 is preferred since we can perform actions immediately when we detect an anomaly. However, as we will see later, there are significant limitations with ROS 2 to make this a reality.

\subsection{A basic AD system}

Our data consists of timestamps for every message passed by each node in the system. A single data point is a sensor-timestamp pair, down to the millisecond. For intuition to drive our AD model, consider that we may define many different metrics or features from the data. Surprising values or combinations of values for any of these metrics could suggest the presence of anomalies:
\begin{itemize}
\item Rate of messages per topic or across all topics
\item Rate of messages per time, day of the week, etc.
\item Autocorrelation of messages: How often does a message of one type tend to follow a message of another type within a particular time-lag window?
\end{itemize}



\section{Results}\label{results}

In order to test our basic AD system, we ran the Turtlebot along specified paths, injecting behaviors such as jerky direction changes, controller disconnects, counter-intuitive path following, and varying speeds. Each path was tested with a random combination of all behavior modifications multiple times. The basic AD system performed reasonably well, capturing anomalies ~70\% of the time. Due to limitations, discussed below, we were unable to verify how the AD system would react after correcting any observed anomalies in the behavior.

While building utilities to capture messages and a real-time AD system, we encountered several limitations with ROS 2, the most pressing of which we enumerate here.
\begin{enumerate}
	\item A node that is solely a Subscriber gets no results from \mintinline{python}{get_topic_names_and_types()} in class Node. In order to get all topic names and types, the node must also be a Publisher.
	\item There are no easily available API methods to access the underlying ROS 2 DDS implementation (eProsima RTPS) that we could find for the Python interface.
	\item Writing messages to disk using the all-subscriber node (Method 1, \ref{our_messages}) causes system hangups and dropped messages in the pub-sub queues. \textbf{This is a significant hurdle to real-time AD systems.}
	\item \verb|rosbag| cannot listen to new topics if they appear after the command to launch \verb|rosbag| is executed. 
	\item There is no simple utility to convert \verb|rosbags| to formats such as JSON, YAML, CSV, and Protobufs. Data scientists may not have the easiest time handling the \verb|rosbag| API, but their toolkits already support the listed formats.
\end{enumerate}

Auxiliary limitations that were faced included the lack of good documentation for many basic ROS 2 tasks, an ever-changing development environment, and hard-to-find community support for the platform. We expect these problems to be solved naturally as ROS 2 matures as a platform. 


\section{Recommendations for ROS 2}

The decentralized (by default) graph structure of a ROS 2 robot sets it apart from its predecessor, as explained in \cite{ROS2Design}. The design of ROS 2 allows for easier development, faster and more understandable inter-node communication dynamics, and a closer alignment to the ideal ``pub-sub" design architecture. However, this decentralization of ROS 2 raises questions about the available goals of an AD system implemented in ROS 2. As described in Section \ref{results}, the current design and implementation of ROS 2 prove to be a hindrance to simple, non-real-time AD applications. In order to be future-facing and to handle the open problems outlined in Section \ref{open_problems}, ROS 2 must evolve and implement some alterations to its design specification.
Our recommendations relate to optimizing the design of ROS 2 to enable advanced and future AD methods to operate most effectively. The basis for our recommendations is a first-principles analysis, as there is not much of an open-source precedent for AD in robotics, and particularly not one for ROS 2.

The following are the core ingredients that we see as necessary for a successful AD system on a robot:
\begin{enumerate}
	\item \textbf{Data:} Sensor data of all kinds are the fundamental inputs to AD algorithms. If real-time AD is needed, real-time data must be collected. If OS temperature is an important indicator of a near-future node malfunction, temperature data must be collected.
	\item \textbf{Data transport/centralization:} Specifically, any node running AD software needs to have data feeds from all other nodes or processes that are relevant for the targeted class of anomalies. This introduces a communication overhead cost.
	\item \textbf{Computational nodes:} Once all data relevant for anomaly detection is being supplied to a central node, the challenge becomes how to derive insight from it. In general, such analysis is computationally costly. To maintain a strong AD capability on a complex robot, ROS needs to support large compute nodes.
\end{enumerate}

Given these core ingredients, we make the following recommendations for ROS 2 based on our findings:

\begin{enumerate}
	\item Introduce strict value ranges into messages (\ref{open_problems_invalid}).
	\begin{itemize}
		\item Messages define what data a listener can accept, but the listener considers only a set range of values valid for any given field. When values out of this range are received, the listener can error out.
		\item Listeners must implement their own range-checking, which is a hurdle and can often be poorly managed.
		\item A message type with built-in range checks can be denied on the publisher's side before the listener ever has to interact with it.
	\end{itemize}
	\item Provide automatic subscription to new topics and messages in \verb|rosbag| (\ref{our_messages}).
	\begin{itemize}
		\item In order to build a real-time AD system, we need to be able to detect when nodes enter and leave the system. In its current form, and its new proposed design \cite{ROS2Design}, \verb|rosbag| cannot detect when a node creates a new topic or an existent one is removed.
		\item Having a utility that can track the dynamism in node publishing behavior is critical, especially during a node crash.
	\end{itemize}
	\item Develop a better profiling environment
	\begin{itemize}
		\item In order to understand the system and to collect valuable diagnostic data, a robust profiling environment needs to exist within the ROS ecosystem.
		\item Currently, many people rely on language specific profiling tools, the \verb|/statistics| topic, and self-created packages to collect system data.
	\end{itemize}
	\item Integrate best-known-state tracking and recovery (\ref{open_problems_fixing})
	\begin{itemize}
		\item There are no ways to backup the state of a node and to restore it to a given point, a method which can quickly resolve many issues with an node without resorting to a complete restart.
		\item System snapshots in general can help with debugging and profiling of performance.
	\end{itemize}
	\item Introduce state introspection (\ref{open_problems_hierarchy}, \ref{open_problems_fixing})
	\begin{itemize}
		\item The ROS graph needs to be known and all states need to be dynamically verified in order to build and maintain a structure of hierarchy.
		\item Knowing the initial (or best) structure and state of all nodes in the ROS system will provide a known target from which to recover from.
	\end{itemize}
	\item Provide a safe mode (\ref{open_problems_fixing})
	\begin{itemize}
		\item When nodes fail, it is essential that core functionalities are still available.
		\item In a safe mode, basic functionalities will be provided to recover the robotic system, such as when in tele-operation.
	\end{itemize}
\end{enumerate}

\section{Conclusion}

Recently, the NTSB and BEA \cite{ntsbTesla2018, af447Report} filed reports that discussed fatal accidents in autonomous systems on land and in air due to improper edge case and anomaly handling by autonomous components in both robotic systems. The reports highlight how these robotic systems were unable to react outside of well understood operating envelopes and unable to alert their operators about their failures in meaningful ways. These reports echos the necessity of understanding the various issues in anomaly detection when deploying a robotic so that spurious or intentional anomalous behavior can be understood and handled before the resulting actions are dangerous both to the robot and the objects around it.

As anomaly detection for robotic systems continues to grow as a point of interest, upcoming robotics platforms need to support these capabilities on a first-class basis. Furthermore, it is clear that robotic anomaly detection is still in its nascent stage. Much work needs to be done to fully understand and handle anomalous behaviors in robotic systems in order to achieve complete, trustworthy autonomy.


\section*{ACKNOWLEDGMENT}

The authors thank Heather Evans, William Shaw, and Hollen Barmer at CMU SEI for their support of this research. Drs. John Dolan and David Bourne at The Robotics Institute at CMU gave us insight into various aspects of robotics.

DISTRIBUTION A. Approved for public release: distribution unlimited.
Copyright 2018 Carnegie Mellon University. All Rights Reserved.
This material is based upon work funded and supported by the Department of Defense under Contract No. FA8702-15-D-0002 with Carnegie Mellon University for the operation of the Software Engineering Institute, a federally funded research and development center.
References herein to any specific commercial product, process, or service by trade name, trade mark, manufacturer, or otherwise, does not necessarily constitute or imply its endorsement, recommendation, or favoring by Carnegie Mellon University or its Software Engineering Institute.
DM18-1034


\bibliography{references_goldmaster}
\bibliographystyle{IEEEtran}

\end{document}